\documentclass{article}
\usepackage[square,sort,comma,numbers]{natbib}
\usepackage{xr}

\RequirePackage[OT1]{fontenc}
\RequirePackage{amsthm,amsmath}
\RequirePackage[numbers]{natbib}

\usepackage{upgreek}
\usepackage{url}
\RequirePackage{graphicx}
\RequirePackage{xcolor,caption,subcaption}
\usepackage[colorlinks=True,citecolor=blue,urlcolor=blue,pagebackref=true,backref=true]
{hyperref}
\renewcommand*{\backrefalt}[4]{%
    \ifcase #1 \footnotesize{(Not cited.)}%
    \or        \footnotesize{(Cited on page~#2.)}%
    \else      \footnotesize{(Cited on pages~#2.)}%
    \fi}

\usepackage{xr}

\makeatletter
\newcommand*{\addFileDependency}[1]{
  \typeout{(#1)}
  \@addtofilelist{#1}
  \IfFileExists{#1}{}{\typeout{No file #1.}}
}
\makeatother

\usepackage{natbib}

\usepackage{graphics}
\usepackage{graphicx}
\usepackage{psfrag}

\usepackage{color}

\usepackage{amssymb,bbm}

\usepackage{nicefrac}

 \usepackage{tabularx}%

\usepackage{enumitem}
\usepackage{booktabs}
\usepackage{caption,subcaption}

\usepackage{mathtools}
\usepackage{authblk}

\newtheorem{theorem}{Theorem}[section]
\newtheorem{lemma}[theorem]{Lemma}
\newtheorem{cor}[theorem]{Corollary}
\theoremstyle{remark}

\newtheorem*{rmk}{Remark}

\newcommand{\mr}{\mathrm}
\newcommand{\mb}{\mathbb}

\newcommand{\mP}{\mathbb{P}}
\newcommand{\R}{\mathbb{R}}

\newcommand{\PP}{\mathbb{P}}

\title{On Posterior Inference for the Number of Clusters in Dirichlet Process Mixture Models}
\author[1]{Chiao-Yu Yang\thanks{chiaoyu@berkeley.edu}}
\author[2]{Eric Xia\thanks{ericzxia@berkeley.edu}}
\author[3]{Nhat Ho\thanks{minhnhat@utexas.edu}}
\author[4]{Michael I. Jordan\thanks{jordan@cs.berkeley.edu}}
\affil[1,2,4]{Department of Statistics, UC Berkeley}
\affil[4]{Department of EECS, UC Berkeley}
\affil[3]{Department of Statistics and Data Sciences, UT Austin}

\date{}
\begin{document}
\maketitle

\begin{abstract}
Dirichlet process mixture models (DPMMs) and Pitman-Yor process mixture models (PYPMMs) have been widely used in density estimation, mode detection, and other nonparametric estimation problems. Due to their discrete, combinatorial nature of the underlying random measures, these models have also been suggested for inference in clustering problems, but it has been shown that these methods are inconsistent in the number of components.  We contribute to this line of work by giving quantitative statements about the clustering behavior. Specifically, we provide nonasymptotic and asymptotic lower bounds on the ratio between the posterior probabilities of consecutive number of clusters in these models under different choices of prior for the parameters.
\end{abstract}

\section{Introduction}
\label{Sec:introduction}
In the wake of the seminal work of Ferguson~\cite{ferguson1973bayesian, Blackwell-MacQueen}, a line of research on Bayesian nonparametric inference emerged that was based on the use of the Dirichlet process and related combinatorial stochastic processes, such as the Pitman-Yor process~\cite{pitman1997two}, as prior distributions. Most commonly, these priors are used in the context of mixture models, where both the prior and the posterior place probability mass on an unbounded number of mixture components~\cite{Antoniak}.

Statistical applications for these models include a variety of problems in density estimation~\cite{escobar1995bayesian, ghosal2010dirichlet, Ghosal-Ghosh-vanderVaart-00, Shen-Wasserman-01, ghosh2003bayesian, Ghosal-vanderVaart-01, Ghosal-vanderVaart-07, Walker-Lijoi-Prunster-07, Ghosal-vanderVaart-07b} and parameter estimation~\cite{Teh-etal-06, Rodriguez-etal-08, Fox-etal-11, Paisley-etal-15}.  Further applications can be found in engineering and scientific fields such as computer vision, information retrieval, economics, astronomy, molecular biology, and genetics \cite{Teh-etal-06,Rodriguez-etal-08,otranto2002nonparametric,escobar1995bayesian,lartillot2004bayesian,huelsenbeck2007inference,medvedovic2002bayesian}. In these latter applications, the problem is often one of clustering, and Bayesian nonparametric mixture models are viewed as a substitute for classical methods such as $K$-means, which generally require the number of components $K$ to be known a priori. Particularly in the setting of dynamically growing data sets, it is difficult and even unreasonable to fix the number of clusters a priori, and Bayesian nonparametric mixture models have been considered as a natural way to circumvent this issue since they allow the number of clusters to be random and subject to posterior inference.

However, there is little theoretical support for this methodology.  Unlike the case of density estimation, results on the convergence of the number of components in Bayesian nonparametric mixtures are largely absent from the literature. Indeed, it has been demonstrated that under certain parametric assumptions, these models exhibit posterior inconsistencies in the number of clusters when the underlying data-generating distribution satisfies some mild conditions~\cite{miller2013simple,miller2014inconsistency}. Moreover, in practice, it has been observed that inference based on Dirichlet process or Pitman-Yor process mixture models can generate small clusters that do not reflect the underlying data-generating process, a serious concern when the real number of components is small.

We study the posterior distribution of the number of clusters of Bayesian clustering models based on the Dirichlet process and the Pitman-Yor process. In contrast to previous work on this topic, we do not assume the underlying data-generating process to be be a specific parametric family or even a finite mixture model, so our results apply to nonparametric data-generating processes as well.  In this general setting, there is no analytic form for the posterior distribution; nonetheless, we have obtained an analytical characterization of the ratio between the probability of obtaining $k+1$ clusters and that of obtaining $k$ clusters, for any positive $k$. We provide novel lower bounds on these ratios, denoted  $R(s|\{x_i\}_{i=1}^n)$, where $x_{1}, \ldots, x_{n}$ are exchangeable observations. We study the setting where the prior on the parameters is either Gaussian or uniform. Our results are as follows:
\begin{itemize}
    \item When the prior on the parameters is uniform over a bounded subset of $\mathbb{R}$, denoted~$\Theta$, the lower bound on $R(s|\{x_i\}_{i=1}^n)$ is of order $\frac{\alpha}{s|\Theta|}$, under a Dirichlet process mixture model, where $\alpha$ is the  scale parameter of the Dirichlet process prior. In the case of a Pitman-Yor process mixture model, the lower bound becomes $\frac{\alpha}{s|\Theta|}+\frac{\xi}{|\Theta|}$ up to a universal constant, where $\xi$ denotes the positive discount parameter. These bounds are nonasymptotic.
    \item When the prior on the parameters is Gaussian, $\mathcal{N}(0, \sigma^2)$, the asymptotic lower bound on $R(s|\{x_i\}_{i=1}^n)$ is of order $\frac{\alpha}{s^2 (1 + \sigma)}$ under a Dirichlet process mixture model and of order $\frac{\alpha}{s^2 (1 + \sigma)}+\frac{\xi}{s(1+\sigma)}$ under a Pitman-Yor process mixture model.
\end{itemize}
These lower bounds provide a fine-grained understanding of the posterior distribution induced by the Dirichlet process and the Pitman-Yor process on the number of clusters. We note, however, that in general there does not exist a finite upper bound. Such an upper bound, should it exist, would require additional assumptions on the data-generating process.

Besides the lower bounds we demonstrate in this paper, we note that one can use our techniques to obtain results for more specific cases where additional assumptions on the data-generating process are imposed.

The remainder of the paper is organized as follows. We begin in
Section~\ref{Sec:prelim} with background on Dirichlet process and Pitman-Yor process mixture models.
Section~\ref{Sec:result_DPMM} is devoted to a study of the posterior distribution of the number of clusters from Dirichlet process mixture models under two choices of priors on the parameters: uniform and Gaussian priors. In
Section~\ref{Sec:PYPMM_result}, we extend these results to the case of Pitman-Yor process mixture models. Section~\ref{Section:experiment} presents the results of experiments that explore some of the consequences of the theoretical results in previous sections. Finally, we conclude the paper with a discussion of potential future work in Section~\ref{Section:Discussion}.

\noindent\textit{Notation:} We use $\{x_i\}_{i=1}^n$ to denote the sample $\{x_1,\cdots,x_n\}$, $[n]$ to denote the set $\{1,\cdots,n\}$, and $A^{n,s}\in\rho_s(n)$ to denote a set $\{A^{n,s}_1,\cdots,A^{n,s}_s\}$ such that all of the elements $A_i$ form an $s$-partition of $[n]$, where $\rho_s(n)$ is the set of all $s$-partitions on $[n]$. For example, $\{\{1,2\},\{3,4\}\}$ is in $\rho_2(4)$.

\section{Preliminaries}
\label{Sec:prelim}
In this section, we provide necessary background on Dirichlet process mixture models and establish key notation for our subsequent analysis of the posterior distribution of the number of clusters in these models.

Dirichlet process mixture models (DPMMs)~\cite{Antoniak, Lo} are specified as follows:
\begin{eqnarray}
    p(A^{n,s}) & : = &\frac{\alpha^s}{\alpha^{(n)}} \prod_{i=1}^s (|A^{n,s}_i|-1)! \label{DPMM1}\\
    p(\theta | A^{n,s}) & : = & \prod_{i=1}^s \pi(\theta_i) \label{DPMM2}\\
    p\big(\{x_i\}_{i=1}^n\big| \{\theta_j\}_{j=1}^s, A^{n,s} \big) & : = & \prod_{j=1}^s \prod_{x_i\in A^{n,s}_j} f_{\theta_j}(x_i), \label{DPMM3}
\end{eqnarray}
where $\pi$ is a prior on the parameter $\theta$ and $\left\{ f_{\theta}(\cdot)\right\}$ is a family of density functions.

In this paper, we focus on the application of DPMMs to clustering problems. The prior for the number of clusters with a data set of size $n$ is as follows:
\[ \mP(K_n=s) = \sum_{A^{n,s}\in \rho_s(n)}p(A^{n,s}).\]
Given this prior distribution, the posterior distribution for the number of clusters admits the following formulation:
\begin{align*}
    & \hspace{-1 em} \mP(K_n=s| \{x_i\}_{i=1}^n) \\
    &= \frac{\mP(\{x_i\}_{i=1}^n |K_n=s)\mP(K_n=s)}{\mP(\{x_i\}_{i=1}^n)}\\
    &\propto \sum_{A\in\rho_s(n)} p(A^{n,s})\cdot \int_{\{\theta_j\}_{j=1}^s} p(\{x_i\}_{i=1}^n | \{\theta_j\}_{j=1}^s)p(\{\theta_j\}_{j=1}^s|A^{n,s})d\{\theta_j\}_{j=1}^s\\
    &= \sum_{A\in\rho_s(n)} p(A^{n,s})\cdot \int_{\{\theta_j\}_{j=1}^s \in \Theta^s} \Big( \prod_{j=1}^s \prod_{x_i\in A_j}f_{\theta_j}(x_i)\prod_{j=1}^s \pi(\theta_j)\Big)d\{\theta_j\}_{j=1}^s.
\end{align*}

To ease the ensuing discussion, we use $m(x_{A^{n,s}_j})$ to denote the probability density function of a cluster of samples, which is given by:
\begin{align*}
m(x_{A^{n,s}_j}) = \int_{\theta_j} f_{\theta_j}(x_{j,1})\cdots f_{\theta_j}(x_{j,a_j})\pi(\theta_j)d\theta_j,
\end{align*}
where $x_{j,1},\cdots,x_{j,a_j}\in A^{n,s}_j$ for all $1 \leq j \leq s$. Given this definition of $m(x_{A^{n,s}_{j}})$, we can rewrite $\mP(K_n = s | x_n)$ as follows:
\begin{align}
    \mP(K_n = s | \{x_i\}_{i=1}^n) & \propto \sum_{A^{n,s}\in \rho_s(n)} \Big( p(A^{n,s})\cdot \prod_{j=1}^s m(x_{A^{n,s}_j})\Big) \nonumber \\
    & = \sum_{A^{n,s}\in \rho_s(n)} \Big(\frac{\alpha^s}{\alpha^{(n)}} \prod_{j=1}^s (|A^{n,s}_j|-1)! \cdot \prod_{j=1}^s m(x_{A^{n,s}_j})\Big).   \label{s_comp_propto}
\end{align}
Early theoretical results on these posterior probabilities can be found in~\cite{miller2013simple,miller2014inconsistency}. A general qualitative result from~\cite{miller2014inconsistency} is as follows.
\begin{theorem}
For a Dirichlet process mixture model, if the component distribution is Gaussian, log normal, Gamma, exponential, Weibull, or if it is discrete and has at least some point with nonzero measure for any choice of parameter, then, provided that the data are independently and identically distributed from a $K^*$-component mixture model, ($K^*$ is bounded), and with probability one we have:
\[ \lim\sup_{n\to\infty}\mP(K_n = K^*| \{x_i\}_{i=1}^n)<1.\]
\end{theorem}
Our goal is to supply a quantitative perspective on these posterior probabilities, in a general setting where we do not confine ourselves to the case of finite or parametric mixtures. In other words, under more general assumptions on the data-generating process, we want to characterize the behavior of the posterior distribution of the number of clusters.

In order to obtain this characterization, we consider the following ratio between its values at $K_{n} = s + 1$ and $K_{n} = s$:
\begin{eqnarray}
    R(s|\{x_i\}_{i=1}^n)
    & := & \frac{\mP(K_n = s+1 \,|\, \{x_i\}_{i=1}^n)}{\mP(K_n = s \,|\, \{x_i\}_{i=1}^n)} \nonumber \\
     & = & \frac{\sum_{A^{n,s+1}\in \rho_{s+1}(n)} \Big( p(A^{n,s+1})\cdot \prod_{j=1}^{s+1} m(x_{A^{n,s}_j})\Big)}{\sum_{A^{n,s}\in \rho_s(n)} \Big( p(A^{n,s})\cdot \prod_{j=1}^s m(x_{A^{n,s}_j})\Big)}. \nonumber
\end{eqnarray}
From the formulation of a DPMM, we further obtain that
\begin{eqnarray}
    & & \hspace{-4 em} R(s|\{x_i\}_{i=1}^n) \nonumber \\
    & = & \alpha\cdot\frac{\sum_{A^{n,s+1}\in \rho_{s+1}(n)} \Big( (\prod_{i=1}^{s+1} (|A^{n,s+1}_i|-1)!)\cdot \prod_{j=1}^{s+1} m(x_{A^{n,s+1}_j})\Big)}{\sum_{A^{n,s}\in \rho_s(n)} \Big( (\prod_{i=1}^s (|A^{n,s}_i|-1)!)\cdot \prod_{j=1}^s m(x_{A^{n,s}_j})\Big)}. \label{ratio}
\end{eqnarray}
Our focus will be to obtain lower bounds on this ratio.
\section{A Quantitative Result for the Posterior of the DPMM}
\label{Sec:result_DPMM}
In this section, we study the posterior distribution on the number of clusters induced by a DPMM. Our first result assumes a uniform prior on a finite interval, with a boundedness assumption on the data. The second result is an asymptotic result for the Gaussian prior.
\subsection{Uniform prior}
\label{sec:Uniform_prior}
We first consider the posterior distribution of the number of clusters of a DPMM when the data lie in a bounded set. Data sets of this form often arise in fields such as biology, genetics, and economics. In this case it is natural to let the parameter space $\Theta$ be a compact set~\cite{Rousseau-2010}.

To ease the complexity of the algebraic aspects of the proof, we specifically consider a simple uniform prior on the parameter space $\Theta$, where $\Theta$ is a bounded segment of $\R$ with size $|\Theta|$. With this choice, we have $\pi(\theta)=1/|\Theta|$ for all $\theta\in\Theta$. We begin with the following result that establishes a lower bound on the ratio $R(s|\{x_i\}_{i=1}^n)$.
\begin{theorem} \label{theorem:lower_bound_ratio_uniform_prior}
Given the DPMM defined in Eq.~\eqref{DPMM3}, with a uniform prior~$\mr{Unif}(\Theta)$ on $\theta$, for sufficiently large $n$, if~$\min(\{x_i\}_{i=1}^n)>\min(\Theta)$ and~$\max(\{x_i\}_{i=1}^n)<\max(\Theta)$, then the ratio~$R(s|\{x_i\}_{i=1}^n)$ between consecutive terms is lower bounded by
  \begin{equation}
    \label{eq:main}
    R(s|\{x_i\}_{i=1}^n) \geq C \cdot \frac{\alpha}{s|\Theta|},
  \end{equation}
where $C > 0$ is a universal constant.
\end{theorem}
\begin{rmk}
Note that in many problems when one applies a uniform prior for the parameters, one expects the uniform prior to have support that is large enough to capture the means of all the components. It is also possible to extend to the case where some samples lie outside $\Theta$ by modifying Lemma \ref{lem:comp1}, but the result would then have to be dependent on the specifics of the distribution and lose generality.
\end{rmk}
\begin{proof}
Given $A^{n,s} \in \rho_{s}(n)$, we define its $(s+1)$-\emph{component extension set} $\eta_{ext}(A^{n,s})$, and its $(s-1)$-\emph{component contraction set} $\eta_{con}(A^{n,s})$ as follows:
\begin{align*}
\eta_{ext}(A^{n,s}) & := \{\tilde{A}^{n,s+1}\in\rho_{s+1}(n):A^{n,s}\in \eta_{con}(\tilde{A}^{n,s+1})\} \\
\eta_{con}(A^{n,s}) & :=\Big\{ \tilde{A}^{n,s-1}\in\rho_{s-1}(n):\, !\exists \, i,j\in[s]:\\
  & \hspace{-1 em} \{A^{n,s}_1,\cdots,A^{n,s}_s,A^{n,s}_i\cup A^{n,s}_j\}\backslash\{A^{n,s}_i,A^{n,s}_j\}=\{\tilde{A}^{n,s-1}_1,\cdots,\tilde{A}^{n,s-1}_{s-1}\}\Big\},
\end{align*}
where we require $s>1$ in the definition of $\eta_{con}(A^{n,s})$.

In words, $\eta_{con}(A^{n,s})$ is the set of partitions in $\rho_{s-1}(n)$ that can be obtained from $A^{n,s}$ by combining two elements of $A^{n,s}$ into one element while keeping everything else the same. On the other hand, $\eta_{ext}(A^{n,s})$ is the set of partitions in $\rho_{s+1}(n)$ for which we can combine two of its elements into one to get $A^{n,s}$. For example, if $s=2$, and we have $A^{n,3}=\{A^{n,3}_1,A^{n,3}_2,A^{n,3}_3\}$, then $\eta_{con}(A^{n,3})$ consists of following three partitions: $\{A^{n,3}_1,A^{n,3}_2\cup A^{n,3}_3\}, \{A^{n,3}_1 \cup A^{n,3}_3,A^{n,3}_2\},\{A^{n,3}_1\cup A^{n,3}_2,A^{n,3}_3\}$, and $\eta_{ext}(A^{n,3})$ is the set of all 4-partitions that can contract to $A^{n,3}$.

We further define the posterior probabilities of a partition to be:
\begin{align*}
p(A^{n,s},x) := p(A^{n,s})\cdot\prod_{i=1}^{s}m(x_{A^{n,s}_i}).
\end{align*}
Given these definitions, we claim that we can rewrite the ratio $R(s|\{x_i\}_{i=1}^n)$ as follows:
\begin{align}
\label{decompose_ratio}
R(s|\{x_i\}_{i=1}^n)&= \frac{\mP(K_n=s+1|\{x_i\}_{i=1}^n)}{\mP(K_n=s|\{x_i\}_{i=1}^n)}\nonumber\\
&=\frac{2}{(s+1)s}\cdot\frac{\sum_{A^{n,s}\in\rho_{s}(n)}\Big(\sum_{A^{n,s+1}\in \eta_{ext}(A^{n,s})}p(A^{n,s+1},x)\Big)}{\sum_{A^{n,s}\in\rho_s(n)}p(A^{n,s},x)},
\end{align}
for $n\geq s+1$.
The proof of this claim is deferred to the end of the proof and we proceed while assuming this result. Note that the last fractional term in the claim takes the generic form $\frac{\sum_a f(a)}{\sum_{a} g(a)}$, which motivates us to study $f(a)/g(a)$ in the following.

For each partition in $\eta_{ext}(A^{n,s})$, there exist exactly two sets that can be combined to get one set in $A^{n,s}$, and all the others are the same as the other clusters of $A^{n,s}$. We now let $\eta_{ext}^{i,j}(A^{n,s})$ denote the subcollection of $\eta_{ext}(A^{n,s})$ such that for any partition in this subcollection, two sets can be combined into $A^{n,s}_i$, and that one set has size $j$. Since the order of sets in each partition does not matter, we will assume for simplicity that the two sets are the $i$-th and $(s+1)$-th clusters. To distinguish the $s$ partition and its induced $s+1$ partitions, we use $B^{n,s}$ to denote the $s$-partition, and $A^{n,s+1}\in \rho_{ext}(B^{n,s})$ to denote its induced partitions. To simplify the notation, we use $b_1,\ldots,b_s$ to denote the cardinalities of $B^{n,s}_1,\cdots,B^{n,s}_s$ and $a_1,\ldots,a_{s+1}$ to denote those of $A^{n,s+1}_1,\ldots,A^{n,s+1}_{s+1}$. Furthermore, in the following derivation, the value $n$ is fixed, so will omit it in the notation unless there is the necessity to discuss different $n$. We obtain the following:
  \begin{align}
      & \hspace{-2 em} \frac{\sum_{A^{s+1}\in\eta_{ext}(B^{s})}p(A^{s+1}|x)}{p(B^{s}|x)} \nonumber \\
      & =\sum_{A^{s+1}\in\eta_{ext}(B^s)}\alpha\cdot\frac{\prod_{i=1}^{s+1}(|A^{s+1}_i|-1)!\,m(x_{A^{s+1}_i})}{\prod_{i=1}^s (|B^s_i|-1)!\,m(x_{B^s_i})} \nonumber \\
      & = \alpha\cdot \sum_{i=1}^s \sum_{j=1}^{b_i-1}\sum_{A^{s+1}\in{\eta}^{i,j}_{ext}(B^s)}\frac{(j-1)!(b_i-j-1)!\,m(x_{A^{s+1}_i})m(x_{A^{s+1}_{s+1}})}{(b_i-1)!\,m(x_{B^s_i})} \nonumber \\
      & = \alpha\cdot \sum_{i=1}^s \sum_{j=1}^{b_i-1}\Big\{\frac{(j-1)!(b_i-j-1)!}{(b_i-1)!}\Big(\sum_{A\in{\eta}^{i,j}_{ext}}\frac{m(X_{A^{s+1}_i})m(X_{A^{s+1}_{s+1}})}{m(X_{B^s_i})}\Big)\Big\}. \label{eq:fixed_s_ratio}
  \end{align}
  Recall that, by definition, $A^{s+1}_i\cup A^{s+1}_{s+1}= B^s_i$.
  \begin{lemma}
  \label{lem:comp1}
   For a set $x_{S}$ with $n_s$ elements and $x_{T}$ with $n_{t}$ elements, suppose $\Theta \subseteq[\theta_{\min},\theta_{\max}]$ is an interval in $\mb{R}$ such that $\min{x_S},\min{x_T} > \theta_{\min}+c$ and $\max{x_S},\max{x_T}<\theta_{\max}-c$ for some $c>0$, we have:
   \[\frac{m(x_S)m(x_T)}{m(x_S\cup x_T)}\geq \frac{c_1\sqrt{2\pi}}{|\Theta|}\cdot\frac{\sqrt{n_s+n_t}}{\sqrt{n_s\,n_t}},\]
   for some positive constant $c_1$ that does not depend on $x_S,x_T$ but only on $c$.
  \end{lemma}

Given Lemma \ref{lem:comp1}, we can derive the following bounds for the term in~\eqref{eq:fixed_s_ratio}:
  \begin{eqnarray}
    \label{eq:ab_ratio}
      \frac{\sum_{A^{s+1}\in\eta_{ext}(B^{s})}p(A^{s+1}|x)}{p(B^{s}x)}
       \geq \frac{\alpha\, c_1\sqrt{2\pi}}{|\Theta|}\cdot \sum_{i=1}^s \sum_{j=1}^{b_i-1}\Big(\frac{b_i}{j(b_i-j)}\Big)^{3/2}.
  \end{eqnarray}
  Note that:
  \begin{align*}
  \sum_{j=1}^{b_i-1}\Big(\frac{b_i}{j(b_i-j)}\Big)^{3/2}\geq \int_{x=1}^{b_i-1}\Big(\frac{b_i}{x(b_i-x)}\Big)^{3/2}dx \geq \frac{4(b_i-2)}{\sqrt{(b_i-1)b_i}}
  \end{align*}
  The inequality implies that the leftmost term is no less than 2 for $b_i \geq 4$. When $b_i=2,3$, simple algebra indicates that $\sum_{j=1}^{b_i-1}\Big(\frac{b_i}{j(b_i-j)}\Big)^{3/2} \geq 2$. Invoking these results, we have the following lower bound:
  \[ \frac{\sum_{A^{s+1}\in\eta_{ext}(B^{s})}p(A^{s+1}|x)}{p(B^{s}|x)} \geq \frac{\alpha \, c_1^2\sqrt{2\pi}}{|\Theta|}\cdot \sum_{i=1}^s 2\cdot I_{b_i\geq 2} \succsim \dfrac{\alpha s}{|\Theta|}. \]
  Combining this lower bound with Eq.~\eqref{decompose_ratio}, we obtain the following evaluation of the ratio between consecutive terms $R(s)$:
  \begin{align}
    R(s|\{x_i\}_{i=1}^n) = \frac{\mP(K_n=s+1|\{x_i\}_{i=1}^n)}{\mP(K_n=s|\{x_i\}_{i=1}^n)} \succsim \dfrac{\alpha}{s|\Theta|}. \nonumber
  \end{align}
As a consequence, we reach the conclusion of the theorem.

\noindent\textit{Proof of claim~\eqref{decompose_ratio}:}
Using equation (\ref{s_comp_propto}), we can rewrite the ratio between the posterior probability of $s+1$ components and that of $s$ components as follows:
    \[ \frac{\mP(K_n=s+1|\{x_i\}_{i=1}^n)}{\mP(K_n=s|\{x_i\}_{i=1}^n)}=\frac{\sum_{A^{s+1}\in\rho_{s+1}(n)}p(A^{s+1},x)}{\sum_{B^s\in\rho_s(n)}p(B^s,x)}.\]
    Note that for each $A^{s+1}\in\rho_{s+1}(n)$, we can merge any two of its $s+1$ parts to obtain some $B^s\in \rho_s(n)$. The number of distinct ways to do so is exactly $\binom{s+1}{2}$. Also, for each $B^s\in\rho_s(n)$, the set $\eta_{ext}(B^s)$ contains all $A^{s+1}\in\rho_{s+1}(n)$ such that they can merge some parts to get $B^s$. Thus, the index of the numerator on the right-hand side counts each $A^{s+1}\in\rho_{s+1}(n)$ exactly $\binom{s+1}{2}$ times, from which the equation follows.
\end{proof}
  \noindent\textit{Proof of Lemma~\ref{lem:comp1}:} For a set $x_S$ with $n_s$ elements, we use the following shorthand:
  \[\overline{x_S} =\frac{\sum_{X\in x_S}X}{n_s};\quad S_{x_S}^2=\sum_{X\in x_S} X^2.\]
  After some algebra, we can verify that
  \begin{eqnarray*}
      & & \frac{m(x_{S})m(x_{T})}{m(x_S\cup x_T)} \\
      & & = \frac{\displaystyle \int_{\theta\in\Theta}\exp \biggr(-\sum_{X\in x_S} \dfrac{(X-\theta)^2}{2}\biggr)d\theta\,\cdot\,\int_{\theta\in\Theta}\exp \biggr(- \sum_{X\in x_T } \dfrac{(X-\theta)^2}{2}\biggr) d\theta}{\displaystyle  |\Theta| \int_{\theta\in\Theta} \exp \biggr(-\sum_{X\in x_S\cup x_T} \dfrac{(X-\theta)^2}{2}\biggr)d\theta}\\
      & & = \dfrac{\frac{2\pi}{\sqrt{ n_s n_{t}}}  \exp \biggr(-\dfrac{(S_{x_S}^2+n_1 \overline{x_S}^2)}{2} - \dfrac{(S_{x_T}^2 + n_t\overline{x_T}^2)}{2} \biggr)P_{x_S}(\Theta) P_{x_S\cup x_T}(\Theta)}{|\Theta| \sqrt{\dfrac{2\pi}{n_s+n_t}} \exp \biggr(-\dfrac{(S_{x_S\cup x_T}^2+(n_s+n_t)\overline{x_S\cup x_T }^2)}{2}\biggr)}\\
      & & = \dfrac{\sqrt{2\pi}}{|\Theta|}\cdot\dfrac{\sqrt{n_s+n_t}}{\sqrt{n_s\,n_t}}\cdot \exp \biggr(\dfrac{n_s \overline{x_S}^2+n_t\overline{x_T}^2-(n_s+n_t)\overline{x_S\cup x_T}^2}{2}\biggr) \\
    & & \hspace{22 em} \times \dfrac{P_{x_S}(\Theta)\,P_{x_T}(\Theta)}{P_{x_S \cup x_T}(\Theta)},
\end{eqnarray*}
  where $P_x(\Theta) := P(\theta\in \Theta | \theta\sim N(\overline{x},\frac{1}{|x|}))$, which can be shown to be at least~$\text{erf}(c|x|/\sqrt{2})$, where $\text{erf}$ is the error function. Hence:
  \[ (\text{erf}(c/\sqrt{2}))^2<\dfrac{P_{x_1}(\Theta)\,P_{x_2}(\Theta)}{P_{x_1 \cup x_2}(\Theta)}<\frac{1}{\text{erf}(c/\sqrt{2})}.\]
  Note that this range in general is very small (around 1) for $c$ that is not too small. For example, the range above is $(0.997^2,1/0.997)$ for $c=3$. Now, we write~$c_1 =(\text{erf}(c/\sqrt{2}))^2$.

  On the other hand,
  \[\frac{n_s\overline{x_S}^2+n_t\overline{x_T}^2-(n_s+n_t)\overline{x_S\cup x_T}^2}{2}=\frac{n_s\, n_t(\overline{x_S}-\overline{x_T})^2}{2(n_s+n_t)}\geq 0.\]
  Combining these results, we obtain the desired inequality. \qedsymbol

The bound in the result of Theorem~\ref{theorem:lower_bound_ratio_uniform_prior} does not require the original distribution to be a mixture distribution. Instead, it holds provided that the true underlying distribution has finite and nonzero variance.

In particular, note that by the law of large numbers, if the observations in $x_S,x_T$ are i.i.d., we may consider the empirical average of the exponential term in Lemma \ref{lem:comp1} and Eq.~(\ref{eq:fixed_s_ratio}):
\[\exp\Big(n_sn_t(\overline{x_S}-\overline{x_T})^2/2(n_s+n_t)\Big).\]
The term inside the exponential is approximately $\frac{\sigma\sqrt{n_sn_t}}{2\sqrt{n_s+n_t}}Z$-distributed as $n_s$ and $n_t$ go to infinity, where $Z$ is the standard normal distribution and $\sigma^2$ is the variance of the original distribution. Using the moment-generating function of the $\chi_2$ distribution, we can see that this term grows unbounded as $n_s$ and $n_t$ grow, in contrast to the constant value one we derived in the proof of the Theorem~\ref{theorem:lower_bound_ratio_uniform_prior}. Therefore, given the result of Theorem~\ref{theorem:lower_bound_ratio_uniform_prior}, we obtain the following corollary:
\begin{cor}
\label{cor:uniform_prior_asymptotics}
If the conditions in Theorem~\ref{theorem:lower_bound_ratio_uniform_prior} hold as $n \to \infty$, then for any $s$ fixed, we have:
\[ \lim_{n\to\infty}R(s|\{x_i\}_{i=1}^n)\to\infty.\]
\end{cor}
We note that the rate at which $R(s|\{x_i\}_{i=1}^n)$ grows is unknown and may be very slow when $\sigma^2$ is small, due to the complicated combinatorial behavior. Combining the results from Theorem~\ref{theorem:lower_bound_ratio_uniform_prior} and Corollary~\ref{cor:uniform_prior_asymptotics}, we can see that for any distribution with nonzero variance, the posterior probability of obtaining $s+1$ clusters will eventually exceed that of obtaining $s$ clusters, and their ratio grows in an unbounded way. An important implication of this result is that, with a larger sample size we will tend to fit a larger number of clusters, and this model ultimately leads to an infinite number of clusters almost surely. However, in the finite sample-size regime, the model's behavior depends more on the variance of the original distribution. In practice, one should rely on the finite bound in Theorem~\ref{theorem:lower_bound_ratio_uniform_prior} unless additional information about the distribution is available.

\subsection{Gaussian Prior}
\label{sec:gaussian_prior}
We now consider the more commonly used Gaussian prior on the parameter $\theta$~\cite{escobar1995bayesian, MacEachern-Muller}. In particular, we choose the prior density to be that of a univariate Gaussian distribution, $\mathcal{N}(0,\sigma^2)$, with fixed variance $\sigma >0$. Given this choice of prior, we have the following asymptotic  lower bound on $R(s|\{x_i\}_{i=1}^n)$:
\begin{theorem} \label{theorem:lower_bound_ratio_Gaussian_prior}
For the DPMM defined in Eq.~\eqref{DPMM3}, with a Gaussian prior $\mathcal{N}(0,\sigma^2)$ on $\theta$, as~$n$ goes to infinity, the ratio $R(s|\{x_i\}_{i=1}^n)$ satisfies the following asymptotic lower bound:
  \begin{equation}
    \label{eq:main_pypmm_Gaussian_prior}
    \lim_{n\to\infty}R(s|\{x_i\}_{i=1}^n) \geq C \cdot \frac{\alpha}{s^2 (1+ \sigma)},
  \end{equation}
 where $C > 0$ is a universal constant.
\end{theorem}
\begin{rmk}
The result of Theorem~\ref{theorem:lower_bound_ratio_Gaussian_prior} holds asymptotically. Its performance in finite samples is unknown and is complex, depending heavily on the original distribution. It would be of interest to characterize the finite-sample behavior of distributions satisfying certain conditions on variance and the true number of components or the true rate of growth in the number of components for an infinite-component distribution induced by processes such as the Dirichlet process.

\end{rmk}
\begin{proof}
We use the notation from the proof of Theorem~\ref{theorem:lower_bound_ratio_uniform_prior} in this proof. As in Eq.~\ref{eq:fixed_s_ratio}, we study the following term: $(m(x_{A^{s+1}_i})m(x_{A^{s+1}_{s+1}}))/m(x_{B^{s}_i})$, where as in Theorem \ref{theorem:lower_bound_ratio_uniform_prior} we assume $B^s$ to be some $s$-cluster formed by merging two clusters in $A^{s+1}$.

\begin{lemma}

\label{lem:comp2}
   For sets $x_S$ with $n_s$ elements and $x_T$ with $n_t$ elements, suppose $\Theta=\R$ and $\pi(\theta)=\frac{1}{\sqrt{2\pi}}\exp(-\frac{\theta^2}{2})$.  We have:
   \[\frac{m(x_S)m(x_T)}{m(x_S\cup x_T)}\geq\sqrt{\frac{1}{2}\cdot\frac{\tau}{1+\tau}\cdot\frac{n_s+n_t}{n_sn_t}},\]
   with probability approaching one as $n_s+n_t$ goes to infinity.

\end{lemma}

Returning to the computation of the ratio $p(A|x)/p(B|x)$, for fixed $s$ and a partition $B \in \rho_s(n)$, we define
\begin{align*}
    U(B) := \{i\in [s]: b_i \geq \frac{n}{s^2}\}.
\end{align*}
For any $i\in U(B)$, since $b_i$ increases as $n$ increases, we may assume that the aforementioned condition that $F(\tau;x_{B_1},x_{B_2})\geq 0$ holds asymptotically for any fixed proportion (less than one) for all the partitions of $B_i$. Then, for sufficiently large $n$ we have:
\begin{eqnarray*}
      & & \hspace{-2 em} \sum_{A^{s+1}\in\eta_{ext}(B^s)}\frac{p(A^{s+1}|x)}{p(B^s|x)} \nonumber \\
      & & = \alpha\cdot \sum_{i=1}^s \sum_{j=1}^{b_i-1}\frac{(j-1)!(b_i-j-1)!}{(b_i-1)!}\Big(\sum_{A^{s+1}\in\eta_{ext}^{i,j}(B^s)}\frac{m(X_{A^{s+1}s_i})m(X_{^{s+1}A_{s+1}})}{m(X_{B^s_i})}\Big) \nonumber \\
      & & \overset{w.h.p.}\geq \frac{\alpha}{2}\frac{\sqrt{\tau}}{\sqrt{1+\tau}}\cdot \sum_{i\in U(B)} \sum_{j=1}^{b_i-1}\frac{(j-1)!(b_i-j-1)!}{(b_i-1)!}\biggr(\sum_{A\in{\eta}_{ext}^{i,j}(B^s)}\sqrt{\frac{b_i}{j(b_i-j)}}\biggr) \nonumber \\
      & & \succsim \frac{\alpha \sqrt{\tau}}{\sqrt{1+\tau}}.
\end{eqnarray*}
Here, the second step follows with high probability by our previous argument, and the last step follows by a similar argument as in the case of uniform prior, except that we have only a constant term instead of an $s$ in the case of uniform prior.  This is true because it is possible to have $|U(A)|\ll s$, so the result can only be bounded by a constant multiple of $\alpha\cdot \frac{\sqrt{\tau}}{\sqrt{1+\tau}}$ without the $s$ factor in the uniform case.

Finally, note that as $n$ goes to infinity, the result holds with probability one. Therefore, we obtain that
\begin{align*}
\lim_{n\to\infty}R(s|\{x_i\}_{i=1}^n) = \lim_{n\to\infty} \frac{\mP(K_n=s+1|\{x_i\}_{i=1}^n)}{\mP(K_n=s|\{x_i\}_{i=1}^n)} \succsim \frac{\alpha}{s^2}\cdot \frac{\sqrt{\tau}}{\sqrt{1+\tau}}.
\end{align*}
As a consequence, we reach the conclusion of the theorem.
\end{proof}

\noindent\textit{Proof of Lemma~\ref{lem:comp2}:} We have:
\begin{align*}
& \frac{m(x_S)m(x_T)}{m(x_S\cup x_T)} \\
& =\frac{1}{\sqrt{\sigma^2}}\sqrt{\frac{(n_s+n_t)+\frac{1}{\sigma^2}}{(n_s+\frac{1}{\sigma^2})(n_t+\frac{1}{\sigma^2})}} \\
& \hspace{6 em} \times \exp \biggr(\frac{1}{2}\biggr(\frac{n_s^2\overline{x_S}^2}{n_s+\frac{1}{\sigma^2}}+\frac{n_t^2\overline{x_T}^2}{n_t+\frac{1}{\sigma^2}} - \frac{(n_s+n_t)^2\overline{x_S\cup x_T}^2}{(n_s+n_t)+\frac{1}{\sigma^2}}\biggr)\biggr). \nonumber
\end{align*}
To simplify the notation, we let $\tau : = \frac{1}{\sigma^2}$ be the precision, and rewrite this expression as:
\[ \sqrt{\tau}\cdot\sqrt{\frac{n_s+n_t+\tau}{(n_s+\tau)(n_t+\tau)}}\exp\Big(\frac{1}{2}\underbrace{\Big(\frac{n_s^2\overline{x_S}^2}{n_s+\tau}+\frac{n_t^2\overline{x_T}^2}{n_t+\tau} - \frac{(n_s+n_t)^2\, \overline{x_S\cup x_T}^2}{(n_s+n_t)+\tau}\Big)}_{F(\tau;x_S,x_T)}\Big).\]

Note that the term $F(\tau;x_S,x_T)$ is nonnegative at zero, since
\[ F(0; x_S,x_T)= \dfrac{n_sn_t}{n_s+n_t}(\overline{x_S}-\overline{x_T})^2 \geq 0. \]
Solving a quadratic function shows that the equation $F(\tau;x_{1},x_{2})=0$ has its positive root at:
\begin{align*}
\begin{cases} \dfrac{1}{4}\Big[\sqrt{8n_sn_t\dfrac{(\overline{x_S}-\overline{x_T})^2}{\overline{x_S}\,\overline{x_T}}+Q^2}+Q\Big],\quad &\text{if }\overline{x_S}\,\overline{x_T}>0\\
\dfrac{1}{4}\Big[\sqrt{8n_sn_t\dfrac{(\overline{x_S}-\overline{x_T})^2}{\overline{x_S}\,\overline{x_T}}+Q^2}+Q\Big]\text{ or }\emptyset,\quad &\text{if }\overline{x_S}\,\overline{x_T}<0\\
\emptyset \quad & \text{if }\overline{x_S}=0, \overline{x_T}\neq 0\text{ or }\overline{x_S}\neq 0, \overline{x_T}= 0\\
\mb{R}^+ \quad &\text{if }\overline{x_S}=\overline{x_T}=0,
\end{cases}
\end{align*}
where $Q := n_t\dfrac{n_t\overline{x_T}}{n_s\overline{x_S}}+n_s\dfrac{n_s\overline{x_S}}{n_t\overline{x_T}}-2(n_s+n_t)$.

If there is no positive root or every positive number is a root, then $F(\tau;x_S,x_T)\geq 0$ for any $\tau>0$. Otherwise, there exists a unique positive root $r_+(x_S,x_T):=\frac{1}{4}\Big[\sqrt{8n_sn_t\frac{(\overline{x}_S-\overline{x}_T)^2}{\overline{x}_S\overline{x}_T}+Q^2}+Q\Big]$, a random variable depending on $n_s,n_t$, whose probability density function concentrates on increasingly larger values as long as one of $n_s,n_t$ goes to infinity. That is, the root grows larger stochastically as $n_s$ and $n_t$ increase. For any fixed $\tau$, as $n_s+n_t$ goes to infinity, it follows that for any fixed $\tau>0$, the probability that $[0,r_+(x_S,x_T)]$ contains $\tau$ approaches one asymptotically, so we have that $F(\tau;x_S,x_T)\geq 0$.

Note that for any positive integers $n_s,n_t$ and nonnegative number $\tau$, we have
\begin{align*}
\frac{n_s+n_t+\tau}{(n_s+\tau)(n_t+\tau)}>\frac{1}{2}\cdot\frac{1}{1+\tau}\cdot\frac{n_s+n_t}{n_sn_t}.
\end{align*}
The result follows.
\qedsymbol
\section{Extension to Pitman-Yor process mixture models}
\label{Sec:PYPMM_result}
In this section, we extend our results to the Pitman-Yor process mixture model, a model that is similar to the Dirichlet process mixture model but allows a faster rate of creation of clusters. Accordingly, we will see that the ratio between the posterior probabilities of consecutive number of clusters is larger than that of the DPMM.

We retain most of the notation used in the previous section unless otherwise specified. We also use the standard Chinese restaurant process nomenclature to describe the dynamics associated with the partitions formed by the Pitman-Yor process, where the items to be partitioned are referred to as ``customers,'' and partions are referred to as ``tables''~\cite{pitman1997two}.  Letting $(\alpha, \xi)$ denote the parameters of the Pitman-Yor process prior, we obtain:
\begin{align}
 \PP(\text{customer } n+1 \text{ joins table } c \,|\, A^{n,s}) = \begin{cases}
    \frac{|c| - \xi}{\alpha + n} & \text{if } c \leq s,\\
    \frac{\alpha + \xi s}{\alpha + n} & \text{otherwise}.
\end{cases}
\end{align}
The idea is that joining an existing table $c$ has a probability of $\frac{|c| - \xi}{\alpha +n}$, a probability proportional to $|c| - \xi$, the number of customers already at the table, discounted by the value $\xi$, and the probability of joining new table is proportional to $\alpha + \xi K_n$. Then, the probability of obtaining $A^{n,s}$ is:
$$ \PP(A^{n,s}) = \frac{\alpha(\alpha+\xi)\cdots(\alpha + \xi(s-1))}{\alpha^{(n)}} \prod_{i=1}^s (1-\xi)(2-\xi) \cdots (|A^{n,s}_i|  - 1 - \xi),$$
where $\alpha^{(n)} = \alpha(\alpha+1)\cdots(\alpha + n - 1)$. Note that if $|A^{n,s}_i|=1$, the corresponding term in the product is $1$. We denote
$$ |c \ominus \xi|! = (1 - \xi)(2 - \xi) \cdots (c - \xi),$$
and define $|0 \ominus \xi|! = 1$.

Overall, we write: $$(A^{n,s}, s) \sim \mr{PYP}(\alpha, \xi),$$
and define the \emph{Pitman-Yor Process Mixture Model} (PYPMM) as follows:
\begin{flalign}
(A^{n,s},s) &\sim \mr{PYP}(\alpha, \xi)\\
p(\theta \, | \, A^{n,s}, s) &= \prod_{i=1}^s \pi(\theta_i)\\
p(\{x_i\}_{i=1}^n \, | \, \{\theta_j\}_{j=1}^s, A^{n,s}, s) &= \prod_{j=1}^s \prod_{x \in A^{n,s}_j} f_{\theta_j}(x_i), \label{PYPMM3}
\end{flalign}
where $\pi$ is a prior on $\theta$ and $f_{\theta_j}$ belongs to a family of (known) densities.
Moreover, analogously with the case of the DPMM, we obtain the following form for the posterior:
\begin{align*}
& \PP(K_n = s | \{x_i\}_{i=1}^n) \\
&= \frac{\PP(\{x_i\}_{i=1}^n \, | \, K_n=s)\PP(K_n = s)}{\PP(\{x_i\}_{i=1}^n)} \\
&\propto \sum_{A^{n,s} \in \rho_s(n)} \PP(A^{n,s}) \cdot \int_{\{\theta_j\}_{j=1}^n} p(\{x_i\}_{i=1}^n \, | \{\theta_j\}_{j=1}^s)p(\{\theta_j\}\, | A^{n,s}) \, d\{\theta_j\}_{j=1}^s \\
&= \sum_{A^{n,s} \in \rho_s(n)} \PP(A^{n,s}) \int_{\{\theta_j\}_{j=1}^s} \left(\prod_{j=1}^s \prod_{x_i \in A^{n,s}_j} f_{\theta_j}(x_i)\right) \prod_{j=1}^s \pi(\theta_j) \, d\{\theta_j\}_{j=1}^s \\
&= \sum_{A^{n,s} \in \rho_s(n)} \PP(A^{n,s}) \prod_{j=1}^s \int_{\theta_j} \left(\prod_{x_i \in A^{n,s}_j} f_{\theta_j}(x_i) \right) \pi(\theta_j) \, d\theta_j \\
&= \sum_{A^{n,s} \in \rho_s(n)} \PP(A^{n,s}) \prod_{j=1}^s m(x_{A^{n,s}_j}).
\end{align*}
Thus, we have
\begin{align*}
& \PP(K_n = s | \{x_i\}_{i=1}^n) \PP(\{x_i\}_{i=1}^n) \\
& = \sum_{A^{n,s} \in \rho_s(n)} \frac{\alpha(\alpha + \xi) \cdots (\alpha + \xi(s-1))}{\alpha^{(n)}} \prod_{i=1}^s |(|A^{n,s}_i| - 1) \ominus \xi|! \cdot m(x_{A^{n,s}_i}).
\end{align*}
Recall the definitions
\begin{align*}
\eta_{ext}(A^{n,s}) & := \{\tilde{A}^{n,s+1}\in\rho_{s+1}(n):A^{n,s}\in \eta_{con}(\tilde{A}^{n,s+1})\}, \\
\eta_{con}(A^{n,s}) & :=\Big\{ \tilde{A}^{n,s-1}\in\rho_{s-1}(n):\, !\exists \, i,j\in[s]:\\
  & \hspace{- 1 em} \{A^{n,s}_1,\cdots,A^{n,s}_s,A^{n,s}_i\cup A^{n,s}_j\}\backslash\{A^{n,s}_i,A^{n,s}_j\}=\{\tilde{A}^{n,s-1}_1,\cdots,\tilde{A}^{n,s-1}_{s-1}\}\Big\},
\end{align*}
and note that $\eta_{s+1}(A)$ is the set of partitions formed by combining $A_i$ and $A_{s+1}$, and $\tilde{\eta}_s(B)$ is all sets in $\rho_{s+1}(n)$ that can be formed by splitting one of $B$ into two clusters (and then making one of them the ``last cluster''). This yields:
\begin{align}
 R(s | \{ x_i \}_{i=1}^n) & = \frac{\PP(K_n = s+1 \, | \, \{x_i\}_{i=1}^n)}{\PP(K_n =s \, | \, \{x_i\}_{i=1}^n)} \nonumber \\
 & = \frac{2}{(s+1)s} \cdot \frac{\sum_{B^{n,s} \in \rho_s(n)}\left(\sum_{A^{n, s-1}\in\eta_{ext}(B^{n,s})} p(A^{n, s-1}, x)\right)}{\sum_{B^{n,s} \in \rho_s(n)} p(B^{n,s}, x)}.
 \end{align}
Thus, continuing with the proof, we have:
\begin{align}
& \frac{\sum_{A^{n, s+1} \in \eta_{ext}(B^{n,s})} p(A^{n,s+1}|x)}{p(B^{n,s}|x)} \nonumber \\
& = \sum_{A^{n,s+1} \in \eta_{ext}(B^{n,s})} (\alpha + \xi s) \frac{\prod_{i=1}^{s+1} |(|A^{n,s+1}_i| - 1) \ominus \xi|! \cdot \prod_{i=1}^{s+1} m(x_{A^{n, s+1}_i})}{\prod_{i=1}^s |(|B^{n,s}_i| - 1) \ominus \xi|! \cdot \prod_{i=1}^s m(x_{B^{n,s}_i})} \nonumber \\
& =(\alpha + \xi s) \sum_{i=1}^s \sum_{j=1}^{b_i - 1} \sum_{A^{n,s+1} \in \eta_{ext}^{i,j}(B^{n,s})} \frac{|(j-1) \ominus \xi|! \cdot |(b_i - j -1) \ominus \xi|!}{|(b_i-1) \ominus \xi|!} \nonumber \\
& \hspace{19 em} \times \frac{m(x_{A^{n,s+1}_i})m(x_{A^{n,s+1}_{s+1}})}{m(x_{B^{n,s}_i})}. \label{PYPMM_general_eq}
\end{align}
Note that we let $b_i = |B_i|$, and let $\eta_{ext}^{i,j}(B^{n,s})$ be defined to be all $A^{n, s+1} \in \eta_{ext}(B^{n,s})$ obtained by splitting $B^{n,s}_i$ into two clusters of size $j$ and $b_i - j - 1$. We now separate into two cases, where the prior is uniform and Gaussian respectively.

\subsection{Uniform prior}
Similar to our analysis with the DPMM, we first consider the behavior of the posterior distribution for the number of clusters when the prior is a uniform distribution.
\begin{theorem} \label{theorem:py_lower_bound_ratio_uniform_prior}
Given the PYPMM defined in Eq.~\eqref{PYPMM3} with a uniform prior $\mr{Unif}(\Theta)$ on $\theta$, when $n$ is sufficiently large, if $\min(\{x_i\}_{i=1}^n)>\min(\Theta)+c$ and $\max(\{x_i\}_{i=1}^n)<\max(\Theta)-c$ for some $c>0$, then the ratio $R(s|\{x_i\}_{i=1}^n)$ between consecutive terms is lower bounded by
  \begin{equation}
    \label{eq:main_pypmm_uniform}
    R(s|\{x_i\}_{i=1}^n) \succsim \frac{(\alpha + \xi s)}{s|\Theta|}.
  \end{equation}
\end{theorem}
\begin{proof}
By Lemma \eqref{lem:comp1} and under the same condition, we have that when $B^{n,s}_i = A^{n, s+1}_i \cup A^{n,s+1}_{s+1}$ and letting $a_k = |A^{n,s+1}_k|$ for $k \in [s+1]$,
$$ \frac{m(x_{A^{n,s+1}_i})m(x_{A^{n, s+1}_{s+1}})}{m(x_{B^{n,s}_i})} \geq \frac{c_1 \sqrt{2\pi}}{|\Theta|} \frac{\sqrt{a_i + a_{s+1}}}{\sqrt{a_ia_{s+1}}}.$$
Now, we obtain that
\begin{align*}
& \frac{\sum_{A^{n,s+1} \in \eta_{ext}(B^{n,s})} p(A^{n,s+1}|x)}{p(B^{n,s}|x)} \\
&\geq (\alpha + \xi s) \sum_{i=1}^s \sum_{j=1}^{b_i - 1} \frac{|(j-1) \ominus \xi|! |(b_i - j - 1) \ominus \xi|!}{|(b_i - 1) \ominus \xi|!} \\
& \hspace{ 16 em} \times \sum_{A^{s+1} \in \eta_{ext}^{i,j}(B^s)} \frac{c_1\sqrt{2\pi}}{|\Theta|} \cdot \frac{\sqrt{b_i}}{\sqrt{j(b_i - j)}} \\
&= (\alpha + \xi s) \frac{c_1 \sqrt{2\pi}}{|\Theta|} \sum_{i=1}^s \sum_{j=1}^{b_i -1} \frac{|(j-1) \ominus \xi|! |(b_i - j - 1) \ominus \xi|!}{|(b_i - 1) \ominus \xi|!} \cdot \binom{b_i}{j} \sqrt{\frac{b_i}{j(b_i - j)}}.
\end{align*}
Note for $j=1$ and $j= b_i -1$ we have that for $b_i \geq 2$
$$\frac{|(j-1) \ominus \xi|! |(b_i - j - 1) \ominus \xi|!}{|(b_i - 1) \ominus \xi|!} \cdot \binom{b_i}{j} \sqrt{\frac{b_i}{j(b_i - j)}} = \frac{b_i}{b_i - 1 - \xi} \sqrt{\frac{b_i}{b_i - 1}} \geq 1.$$
Plugging this back into the expression above
$$ \frac{\sum_{A^{n,s+1} \in \eta_{ext}(B^{n,s})} p(A^{n,s+1}|x)}{p(B^{n,s}|x)} \geq (\alpha + \xi s) \frac{(0.997)^2 \sqrt{2\pi}}{|\Theta|} \sum_{i=1}^s 2 \cdot I_{b_i \geq 2} \gtrsim \frac{(\alpha + \xi s)s}{|\Theta|}.$$
Then, using Eq.~\ref{PYPMM_general_eq} we get that
\begin{align*}
R(s|\{x_i\}_{i=1}^n) & = \frac{2}{(s+1)s} \cdot \frac{\sum_{B^{n,s} \in \rho_s(n)}\left(\sum_{A^{n,s+1} \in \eta_{ext}(B^{n,s})} p(A^{n,s+1}|x)\right)}{\sum_{B^{n,s} \in \rho_s(n)} p(B^{n,s}|x)} \\
& \geq \frac{2}{(s+1)s} \min_{B^{n,s} \in \rho_s(n)} \sum_{A^{n,s+1} \in \eta_{ext}(B^{n,s})} \frac{p(A^{n,s+1}|x)}{p(B^{n,s}|x)} \\
& \gtrsim \frac{2}{(s+1)s} \frac{(\alpha + \xi s)s}{|\Theta|} \gtrsim \frac{(\alpha + \xi s)}{s|\Theta|} = \frac{\alpha}{s|\Theta|} + \frac{\xi}{|\Theta|}.
\end{align*}
As a consequence, we obtain the conclusion of the theorem.
\end{proof}

\subsection{Gaussian Prior}
We now turn to the case where the prior is a Gaussian distribution. We make similar assumptions on the prior as we made in Section~\ref{sec:gaussian_prior} with the DPMM.
\begin{theorem} \label{theorem:py_lower_bound_ratio_Gaussian_prior}
For the PYPMM defined in Eq.~\eqref{PYPMM3}, with a Gaussian prior $\mathcal{N}(0,\sigma^2)$ on $\theta$, as~$n$ goes to infinity, the ratio $R(s|\{x_i\}_{i=1}^n)$ satisfies the following asymptotic lower bound:
  \begin{equation}
    \label{eq:main_dpmm_Gaussian_prior}
    \lim_{n\to\infty}R(s|\{x_i\}_{i=1}^n)\geq \frac{C (\alpha + \xi s)}{s^2}\cdot \frac{1}{1+\sqrt{\sigma^2}},
  \end{equation}
 where $C > 0$ is a universal constant.
\end{theorem}
\begin{proof}
Most steps are the same as in the derivation for the DPMM. We denote
$$D_{j, b_i} =  \frac{|(j-1) \ominus \xi|! |(b_i - j - 1) \ominus \xi|!}{|(b_i - 1) \ominus \xi|!}.$$
Then, by equation~\eqref{PYPMM_general_eq} we have that
\begin{align*}
 & \sum_{A^{n,s+1} \in \eta_{ext}(B^{n,s})} \frac{p(A^{n,s+1}|x)}{p(B^{n,s}|x)} \\
 & = (\alpha + \xi s) \sum_{i=1}^s \sum_{j=1}^{b_i - 1}D_{j, b_i}\sum_{A^{n,s+1} \in \eta_{ext}^{i,j}(B^{n,s})} \frac{m(x_{A^{n,s+1}_i})m(x_{A^{n,s+1}_{s+1}})}{m(x_{B^{n,s}_i})}\\
 &\overset{w.h.p.}\geq C_0(\alpha + \xi s) \sqrt{\tau} \sum_{i \in U(B^{n,s})} \sum_{j=1}^{b_i - 1} D_{j, b_i} \times \left(\sum_{A^{n,s+1} \in \eta_{ext}^{i,j}(B^{n,s})} \sqrt{\frac{b_i + \tau}{(j + \tau)(b_i - j + \tau)}}\right)
\end{align*}
\begin{align*}
 &\geq C_0(\alpha + \xi s) \sqrt{\tau} \sum_{i\in U(B^{n,s})} \sum_{j=1}^{b_i - 1} D_{j, b_i} \times \left(\sum_{A \in \eta_{ext}^{i,s}(B^{n,s})} \frac{1}{\sqrt{2(1+\tau)}} \sqrt{\frac{b_i}{j(b_i - j)}} \right) \\
 &\geq \frac{C\sqrt{\tau}}{1+\sqrt{\tau}}(\alpha + \xi s),
 \end{align*}
where most of the steps follow similarly as in the proof of the DPMM in Theorem~\ref{theorem:lower_bound_ratio_Gaussian_prior}. Thus we conclude that
 $$ \lim_{n\to\infty} R(s\,|\, \{x_i\}_{i=1}^n) \gtrsim \frac{(\alpha + \xi s)}{s^2} \cdot \frac{\sqrt{\tau}}{1 + \sqrt{\tau}} = \frac{\sqrt{\tau}}{1 + \sqrt{\tau}}\left(\frac{\alpha}{s^2}   + \frac{\xi}{s} \right).$$
As a consequence, we reach the conclusion of the theorem.
\end{proof}
\begin{rmk}
Compared to the DPMM, the PYPMM has a faster rate of table generation, and accordingly a longer tail for the number of clusters. As we can see in Theorems~\ref{theorem:py_lower_bound_ratio_uniform_prior} and~\ref{theorem:py_lower_bound_ratio_Gaussian_prior}, this leads to a larger lower bound in the ratio between consecutive posterior probabilities. Essentially, the lower bound for the uniform case is raised to a constant, and that for the Gaussian case is raised to the order of $1/s$, which corresponds to our intuition that PYPMM is more likely to fit a bigger number of clusters in the posterior.
\end{rmk}
 \section{Experiments}
 \label{Section:experiment}

 In the current section, we discuss the results of numerical experiments that aim to empirically strengthen our understanding of the clustering behavior of the DPMM and PYPMM. Throughout this section, for simplicity of implementation, we consider the Dirichlet mixture of standard normals; i.e., we assume $f_\theta(x) = \frac{1}{\sqrt{2\pi}}e^{-(x-\theta)^2/2\sigma^2}$, where $f_\theta(x)$ is the likelihood function in Eq.~\eqref{DPMM3}. As for the prior density on the parameters, we choose $\pi(\theta)$ to be Gaussian.

To compute the posterior density given the data points, we note that if data points $\{ x_i \}_{i=1}^k$ fall within the cluster, then it follows that
\begin{align*}
p(\theta | \{ x_i \}_{i=1}^n) &\propto p(\theta) p(\{ x_i \}_{i=1}^n|\theta) \\
&= \frac{1}{\sqrt{2\pi}} e^{-\theta^2/2\sigma^2} \prod_{i=1}^n \frac{1}{\sqrt{2\pi}} e^{-\frac{1}{2}(x_i - \theta)^2} \\
&\sim \mathcal{N}\left(\frac{\sigma^2\sum_{i=1}^n x_i}{n\sigma^2 + 1}, \frac{\sigma^2}{n\sigma^2 + 1}\right).
\end{align*}
Moreover, to compute the marginal density of a specific $x$, we compute as follows:
\begin{align*}
f(x) &= \int f_\theta(x) \pi(\theta) \, d\theta \\
&= \int \frac{1}{\sqrt{2\pi}} e^{-\frac{1}{2}(x - \theta)^2} \cdot \frac{1}{\sqrt{2\pi\sigma^2}} e^{-\frac{\theta^2}{2\sigma^2}} \, d\theta\\
&\sim \mathcal{N}(0, \sigma^2 + 1).
\end{align*}

We pick $\alpha = 1$ and $\sigma = 1$ for our simulations. Specifically, we use the ``dirichletprocess'' R package~\cite{dirichletprocesspackage} for these simulations, where the software package runs a Gibbs sampler on the DPMM model. For all the experiments, we run the Gibbs sampler with a burn-in period with $2\times 10^5$ burn-in steps, and then take $10000$ succeeding samples with step size 100.

\subsection{Finite-cluster models}
In this subsection we consider two finite-cluster distributions.  In the first experiment, we generate 300 i.i.d.\ data points following a standard normal distribution. In the second experiment, we generate another 300 data points with the following two-cluster distribution:
\[ f(x)=\frac{1}{2}, \quad \forall x\in [0,1] \cup [2,3].\]

After running the Gibbs sampler, we compute the posterior frequency of the number of clusters $F_s$ for each $s$ with some nonzero count of posterior frequency. This gives an empirical distribution of $\mP(K_n=s |\{x_i\}_{i=1}^n)$, from which we may approximate $R(s|\{x_i\}_{i=1}^n)$. Note, however, that this approximation may be of low quality for large $s$ given the corresponding low posterior frequency.

\begin{figure}[!t]
    \centering
    \begin{subfigure}[t]{0.5\textwidth}
        \centering
        \includegraphics[width=\textwidth]{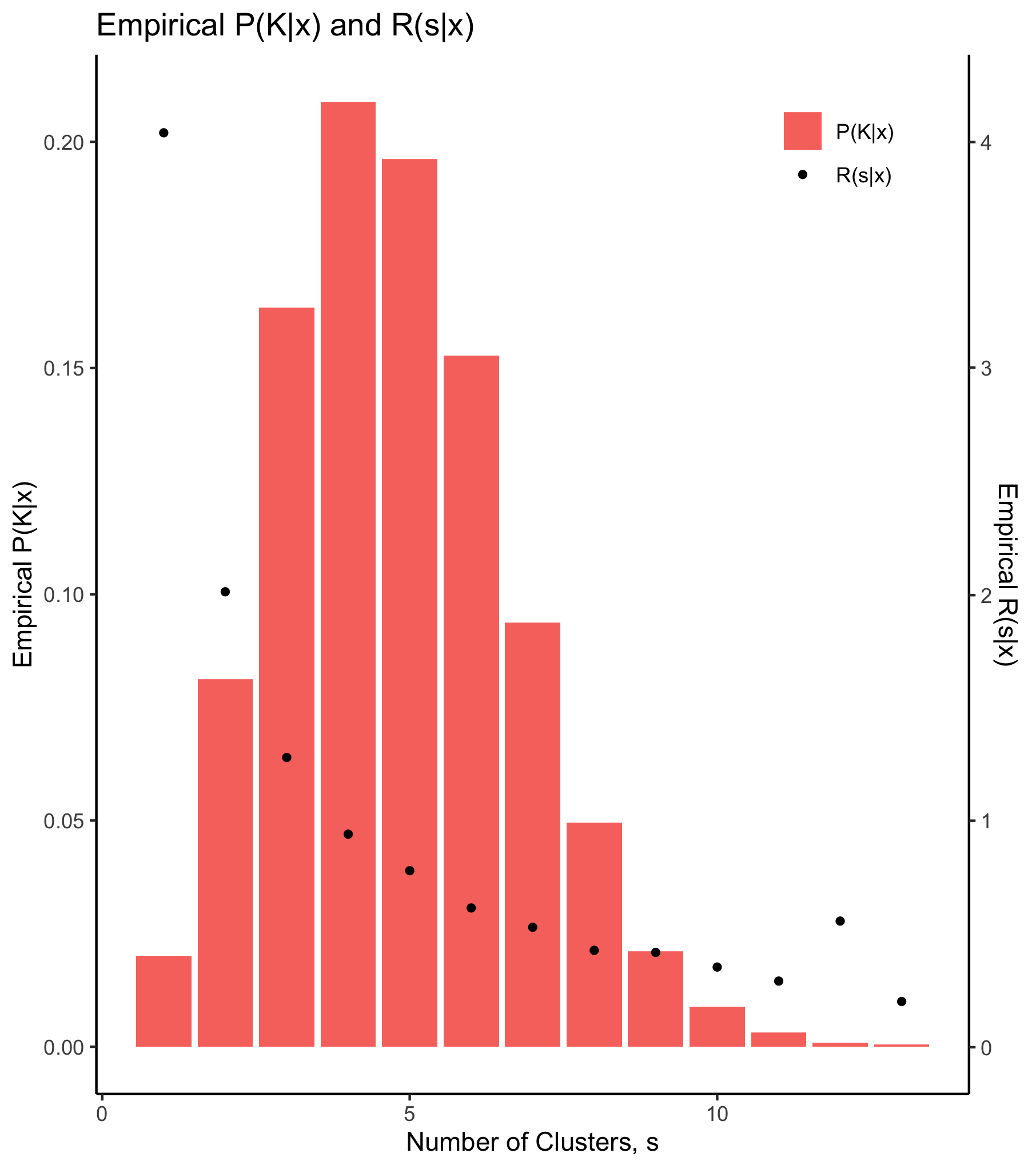}
        \caption{Gaussian, one cluster}
    \end{subfigure}%
    \begin{subfigure}[t]{0.5\textwidth}
        \centering
        \includegraphics[width=\textwidth]{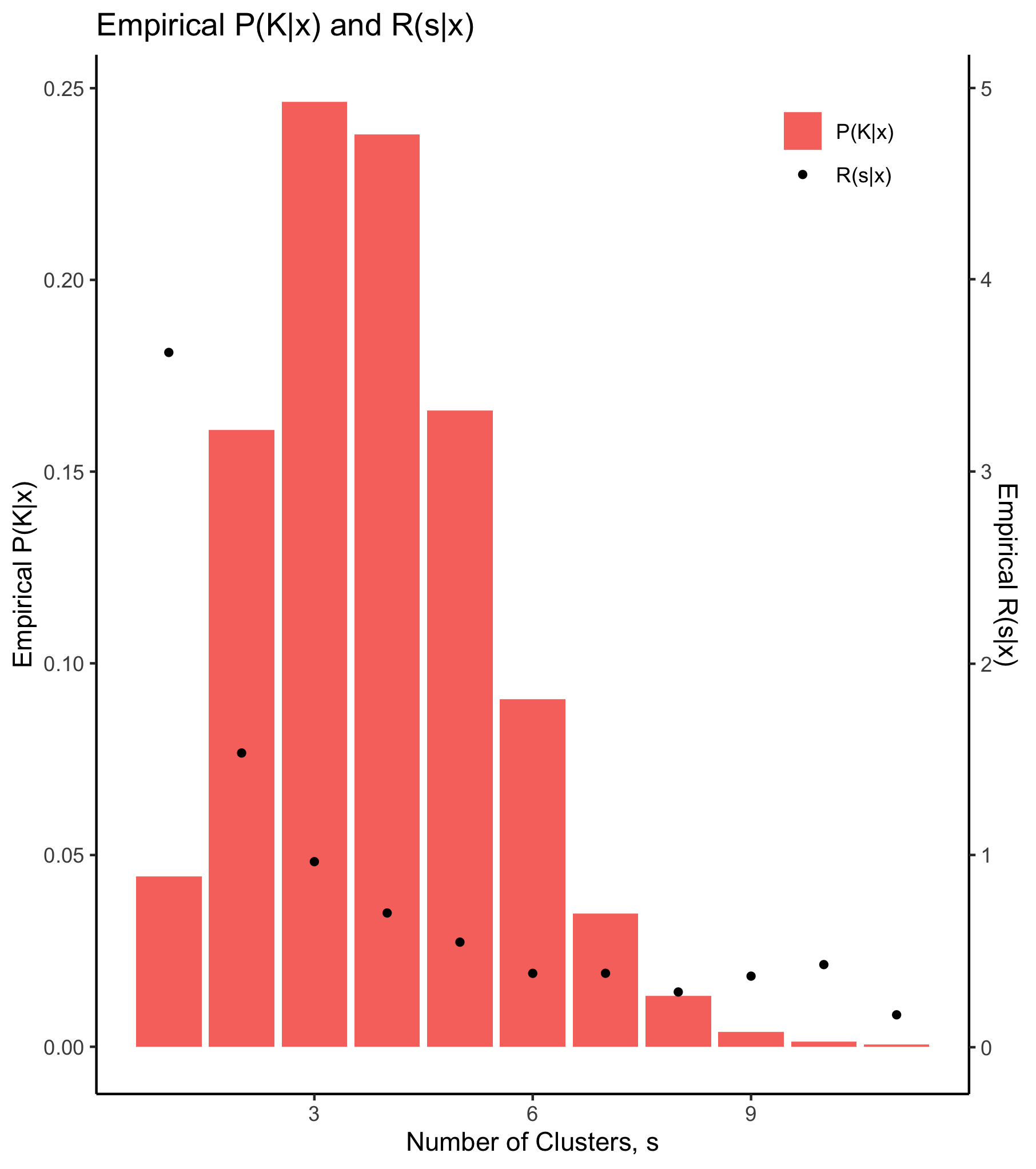}
        \caption{Uniform, two clusters}
    \end{subfigure}%
    \caption{Plots for empirical $\PP(K_n|x)$ and $R(s|x)$}
    \label{fig:finite1}
\end{figure}

Figure \ref{fig:finite1} presents a plot of posterior frequencies and their ratios. The red bars correspond to the posterior frequency of the number of components, and the black dots are the ratios between the posterior frequencies, $F_{s+1}/F_s$.

\begin{figure}
    \centering
    \begin{subfigure}[t]{0.38\textwidth}
        \centering
        \includegraphics[width=\textwidth]{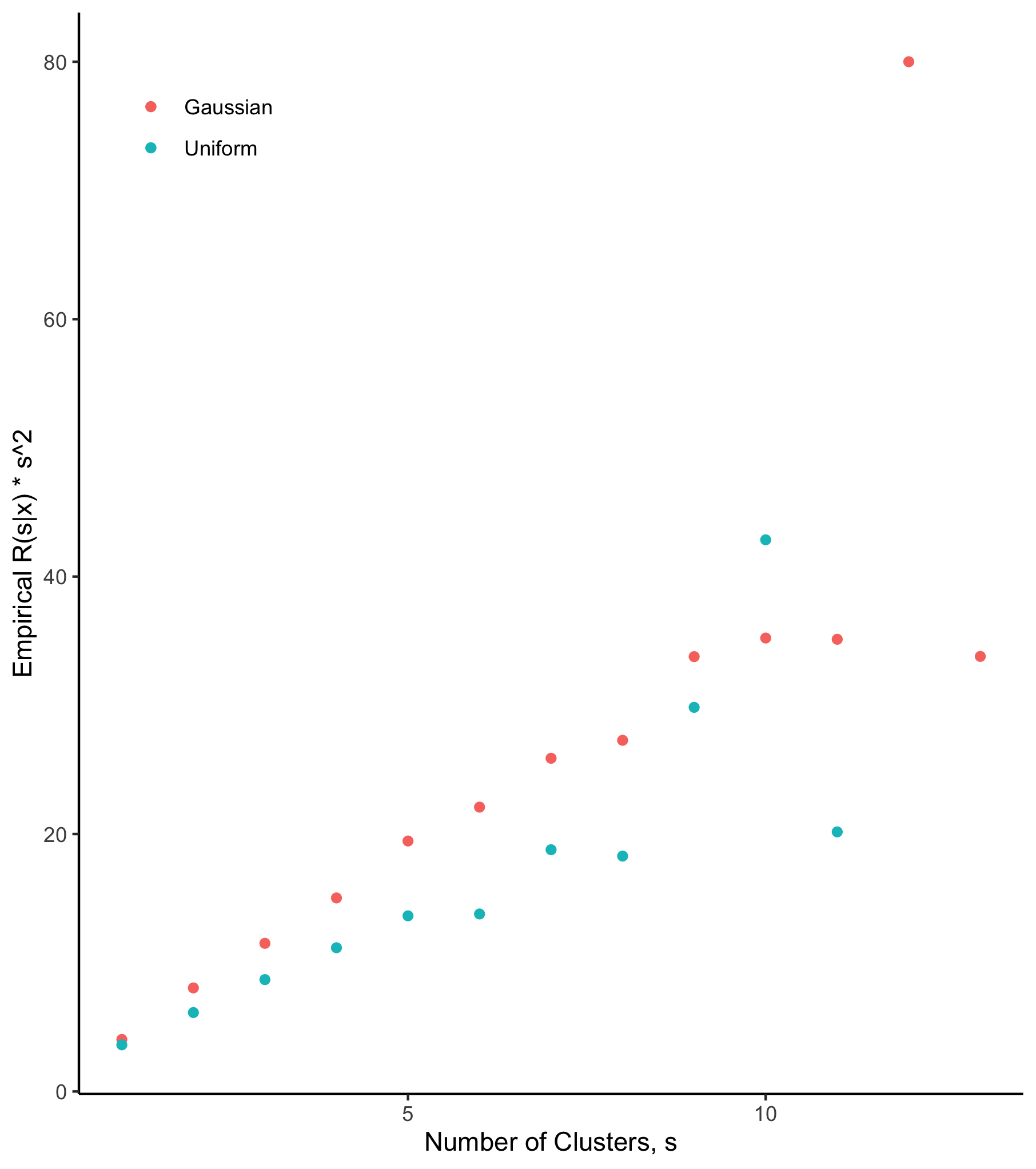}
        \caption{Empirical $\widehat{R}(s|x) * s^2$}
        \label{fig:finite2a}
    \end{subfigure}%
    ~
    \begin{subfigure}[t]{0.38\textwidth}
        \centering
        \includegraphics[width=\textwidth]{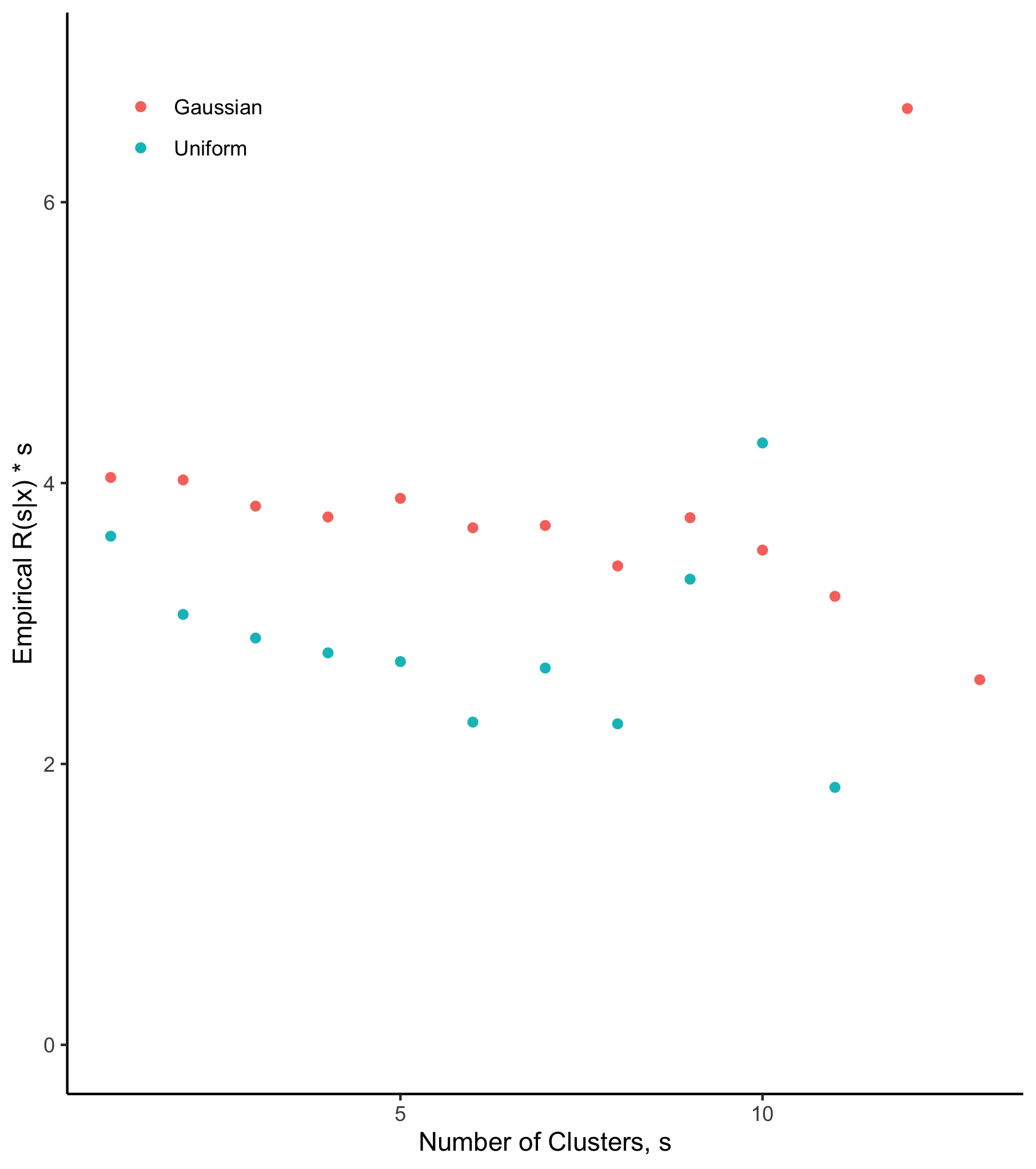}
        \caption{Empirical $\widehat{R}(s|x) * s$}
        \label{fig:finite2b}
    \end{subfigure}%
    \caption{Plots of $\widehat{R}(s|x)$ with respect to different order of powers}
    \label{fig:finite2}
\end{figure}

The theoretical bound asserts that the ratio between consecutive posterior frequencies is at least of the order of $1/s^2$ under a Gaussian prior. To evaluate this, we plot $R(s)\cdot s^2$ versus $s$, which, roughly speaking, should look flat or have an upwards trend. In Figure \ref{fig:finite2a}, we plot $\widehat{R}(s|x)\cdot s^2$ for both the one-cluster Gaussian and two-cluster uniform cases, where we observe the expected upward trend. In Figure \ref{fig:finite2b}, we plot $\widehat{R}(s|x)\cdot s$ to assess whether the relationship is roughly linear. If linearity holds true then we would expect to see the dots for both settings to be roughly flat. Finally we note that it is reasonable to have erratic behavior of $\widehat{R}(s|x)$ when $s$ is large since the posterior frequencies for large $s$ are all very close to zero and the estimator $\widehat{R}$ is unstable in that case.

\subsection{Dirichlet process mixture models}
It is also of interest to investigate the behavior when the data follow an infinite-component distribution.

To study this case, we investigate a Dirichlet process data-generating model, with the $\alpha = 3$, $\theta_i\overset{i.i.d.}\sim N(0, 25)$ for each $i$, and
$$x_i |\theta_j, i\in A_{j}^{n,s}\sim N(\theta_j, 1).$$
Again, we generate 300 samples for three independent experiments, which result in 11, 14, and 17 clusters respectively.

\begin{figure*}[h]
    \centering
    \begin{subfigure}[t]{0.33\textwidth}
        \centering
        \includegraphics[width=\textwidth]{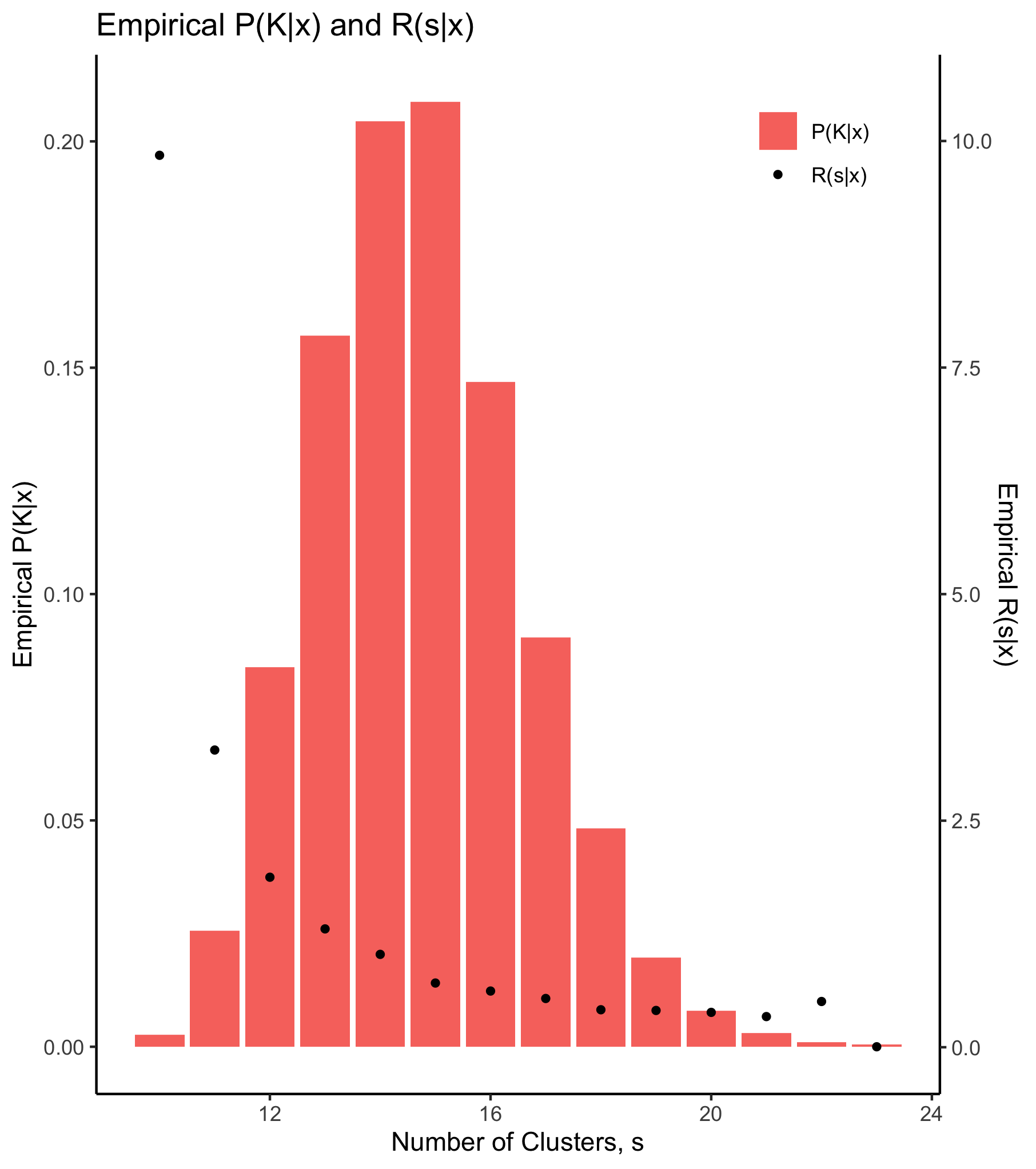}
        \caption{11 clusters}
    \end{subfigure}%
    ~
    \begin{subfigure}[t]{0.33\textwidth}
        \centering
        \includegraphics[width=\textwidth]{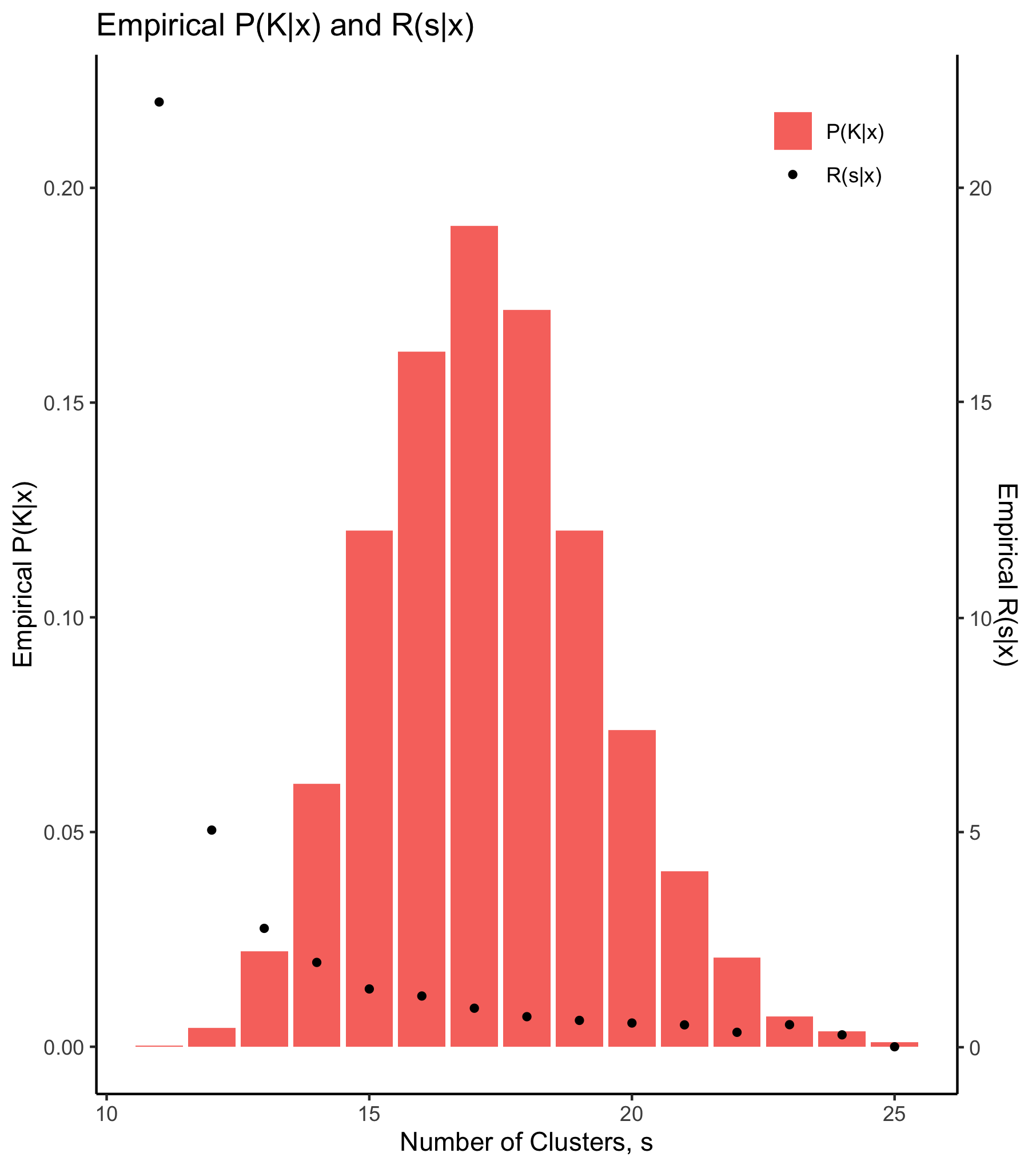}
        \caption{14 clusters}
    \end{subfigure}%
    ~
    \begin{subfigure}[t]{0.33\textwidth}
        \centering
        \includegraphics[width=\textwidth]{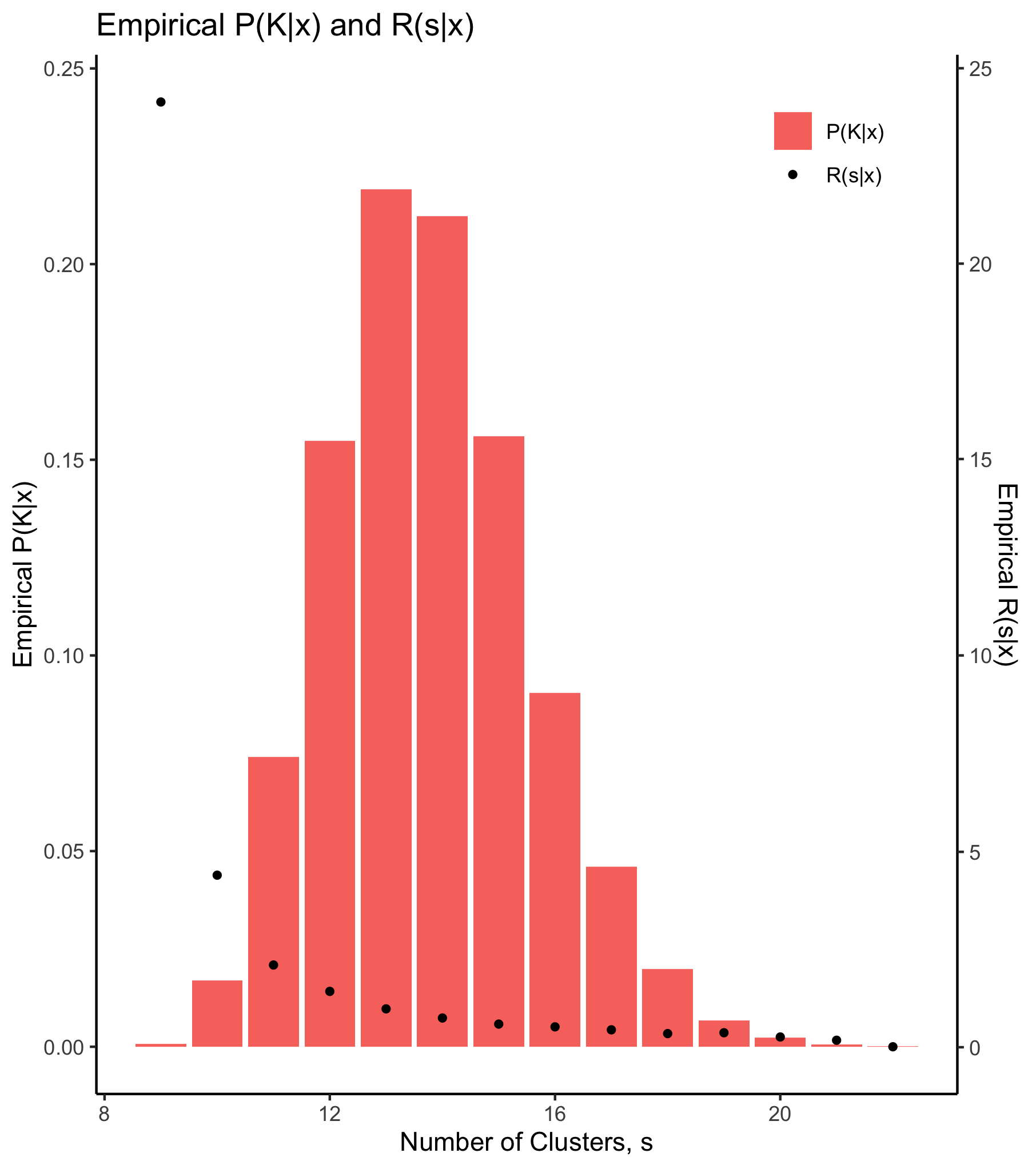}
        \caption{17 clusters}
    \end{subfigure}%
    \caption{Plots for empirical $\PP(K_n|x)$ and $R(s|x)$ in three simulations from Dirichlet process.}
    \label{fig:dp1}
\end{figure*}

\begin{figure*}[!t]
    \centering
    \begin{subfigure}[t]{0.38\textwidth}
        \centering
        \includegraphics[width=\textwidth]{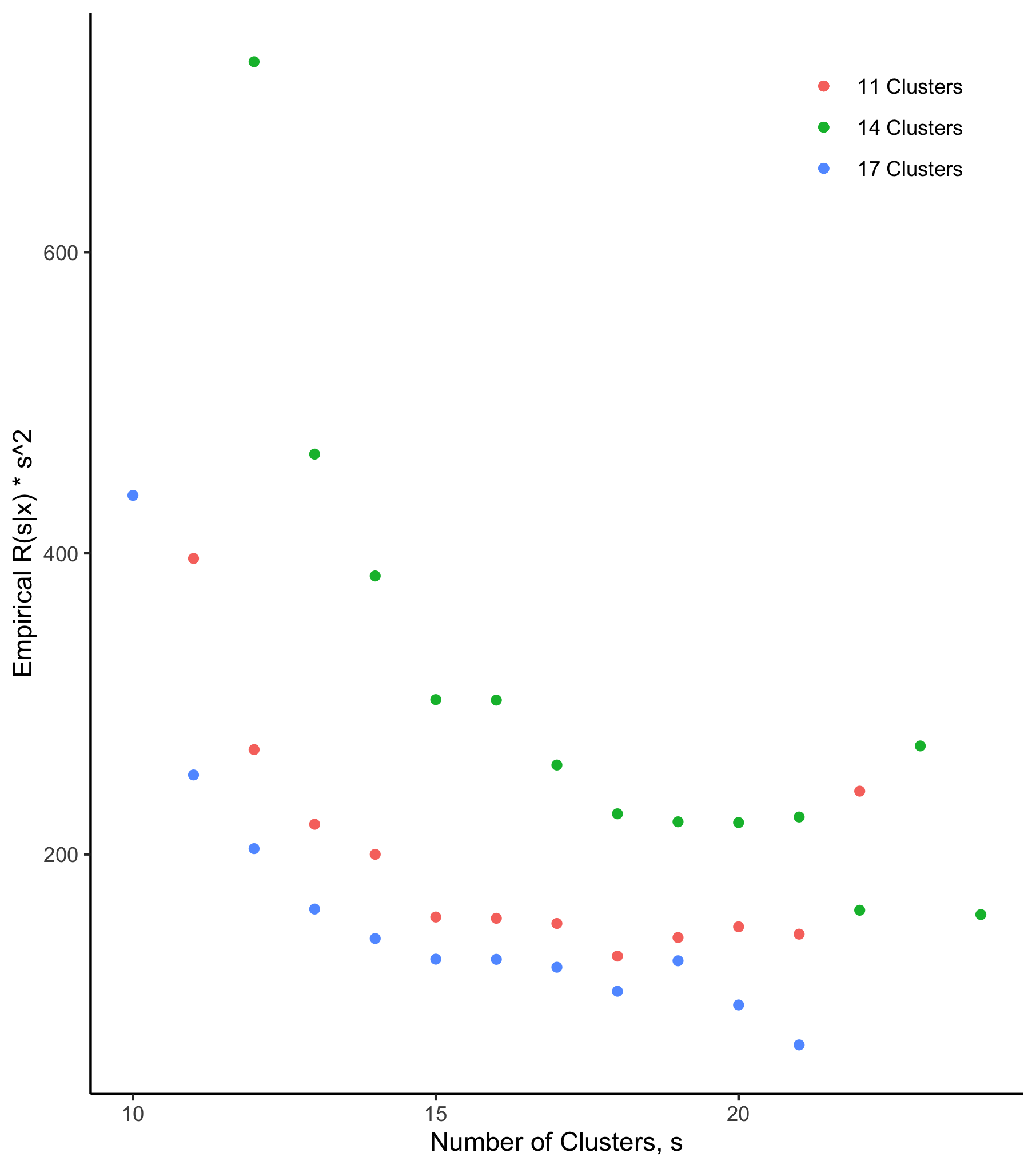}
        \caption{Empirical $R(s|x) * s^2$}
        \label{fig:dp2a}
    \end{subfigure}%
    ~
    \begin{subfigure}[t]{0.38\textwidth}
        \centering
        \includegraphics[width=\textwidth]{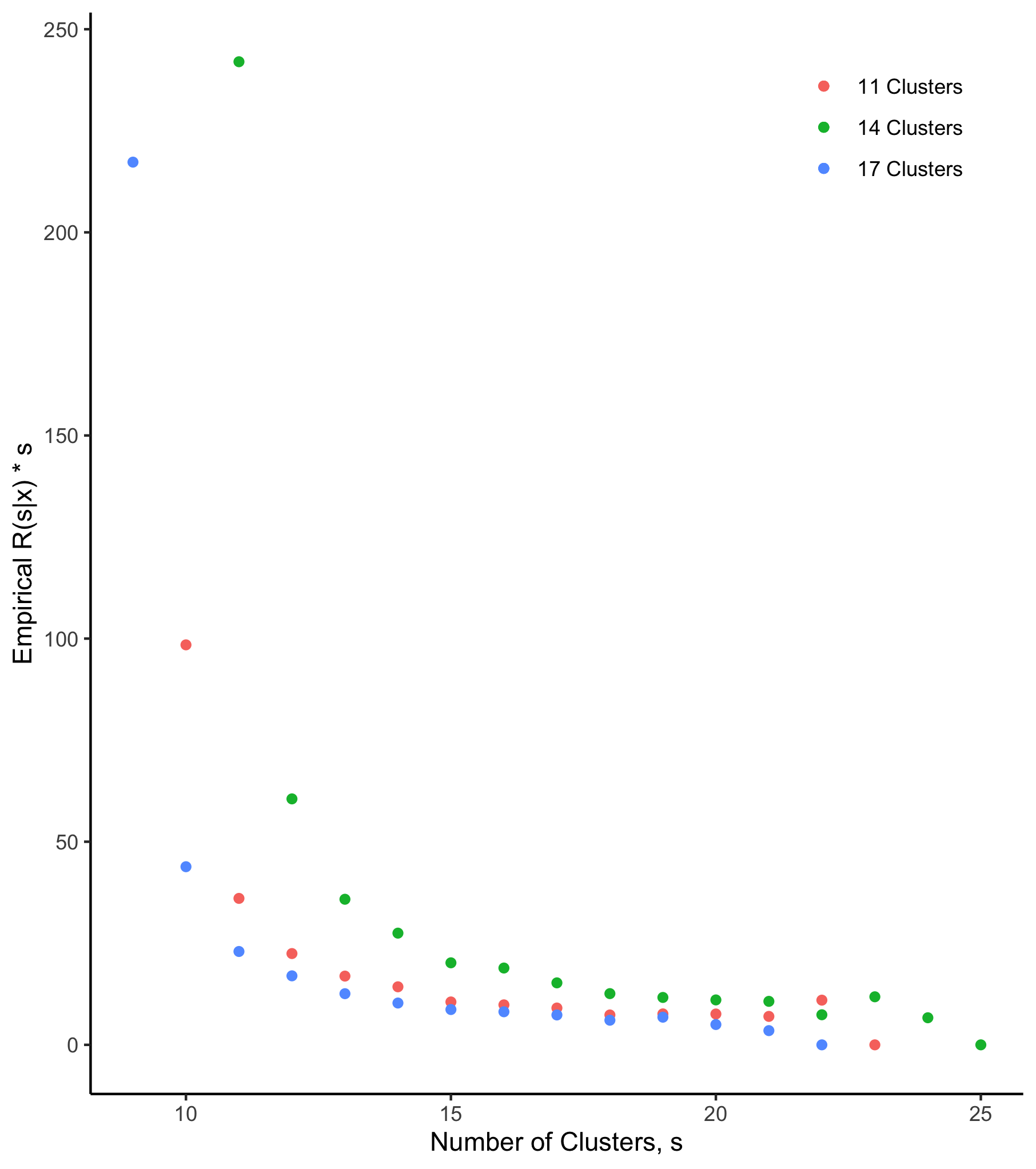}
        \caption{Empirical $R(s|x) * s$}
        \label{fig:dp2b}
    \end{subfigure}%
    \caption{Plots of $\widehat{R}(s|x)$ with respect to different order of powers in three simulations from the Dirichlet process.}
    \label{fig:dp2}
\end{figure*}

We again plot the empirical values of $\PP(K_n|x)$ and $R(s|x)$ in Figure~\ref{fig:dp1}, and investigate the nature of the growth rate of the posterior number of clusters in Figure~\ref{fig:dp2}.  As we see in the latter figure, the ratios between consecutive posterior frequencies of the number of clusters are at least of order $1/s^2$.
\section{Discussion}
\label{Section:Discussion}
We have established lower bounds on the ratio of posterior probabilities $R(s|\{x_i\}_{i=1}^n)$ for the number of clusters for both the Dirichlet process mixture model and Pitman-Yor process mixture model, under several choices of prior distributions on the parameter space.  While the complicated combinatorial structure of the DPMM and PYPMM preclude general characterizations of the posterior distribution, we have shown that it is possible to obtain   analytical results for the posterior distribution of the number of clusters.  We obtained both asymptotic and nonasymptotic results for the distribution of this important quantity.

An interesting open problem is to consider whether our rates are optimal or can be further improved.  In particular, in the case of a Gaussian prior, our simulation suggests that instead of the $1/s^2$ rate predicted by our theory, a tighter $1/s$ rate may be possible. This seems plausible given that the $1/s$ rate arises in the case of uniform prior.

Our results provide a strong negative response to the question of whether basic nonparametric Bayesian models such as the DPMM or the PYPMM are able to infer the true number of clusters when the true number of clusters is finite.  Indeed, our results provide a quantitative refutation of this naive hope.  Additional mechanisms, such as truncation or some form of regularization, will be necessary to obtain consistent inference of the number of clusters in the finite setting. Indeed, recent work has shown that truncation can yield consistency with the number of clusters when the true data-generation mechanism is a finite mixtures~\cite{Ho_truncate}. However, the truncation method brings additional tuning parameters into the picture, including notions of separation of clusters, which may not be easily determined in practice.  The problem of consistent estimation of the number of clusters remains open.

There is, however, another interesting open problem, which arises when the true distribution contains an infinite number of components.  Do the DPMM or the PYPMM guarantee an infinite number of clusters in this case?  Does the rate of growth of clusters in the posterior match that of the data-generating distribution, at least asymptotically?  What about the case in which the truth is Dirichlet process mixture of normals or simply a general Dirichlet process? Does DPMM generate a posterior number of clusters at the same rate as implied by the Dirichlet process? In answering these questions, it will be necessary to derive some sort of upper bound on the $R(s|X)$, and focus on the asymptotic regime.

\section{Acknowledgements}
\label{sec:acknowledge}

We wish to acknowledge support from the by Army Research Office grant W911NF-17-1-0304 and the Mathematical Data Science program of the Office of Naval Research under grant number N00014-18-1-2764.

\bibliographystyle{plain}
\bibliography{main}

\end{document}